\documentclass[runningheads]{llncs}

\usepackage{eccv}

\usepackage{eccvabbrv}

\usepackage{graphicx}
\usepackage{booktabs}
\usepackage{multirow}

\usepackage[accsupp]{axessibility}  

\usepackage{hyperref}

\usepackage{orcidlink}

\begin{document}

\title{CoGoal3D: Collaborative 3D Object Detection with 3D-Aware Fusion and Refinement} 

\titlerunning{CoGoal3D: Collaborative 3D Object Detection with 3D-Aware Fusion}

%
%

\author{
	Zhihao Yang\inst{1} \and
	Zhiyu Xiang\inst{1,2}\thanks{Corresponding author.} \and
	Peng Xu\inst{1} \and
	Tianyu Pu\inst{1} \and
	Kai Wang\inst{1} \and
	Eryun Liu\inst{1} \and
	Dongping Zhang\inst{3} \and
	Yong Ding\inst{1}
}

\authorrunning{Z.~Yang et al.}

\institute{
	Zhejiang University, Hangzhou, China\\
	\email{\{yangzhihao, xiangzy, xxxupeng, 3190105835, kai-wang, eryunliu, dingyong09\}@zju.edu.cn}
	\and
	Zhejiang Provincial Key Laboratory of Multi-Modal Communication Networks and Intelligent Information Processing, Hangzhou, China
	\and
	China Jiliang University, Hangzhou, China\\
	\email{06a0303103@cjlu.edu.cn}
}

\maketitle

\begin{abstract}
V2X collaborative object detection features overcoming the limitations of single-vehicle systems by aggregating environmental features from multiple collaborative agents. However, existing mainstream V2X perception methods mainly focus on 2D BEV object detection. When 3D  detection task is concerned, inferior results are obtained because they ignore the 3D spatial misalignment caused by differing height and attitude among the collaborators. In this paper, we propose a novel collaborative 3D object detection framework called CoGoal3D, which extracts and refines the 3D feature gradually in a two-stage pipeline. In the first stage, a multiscale 3D-aware global fusion module is designed to mitigate the 3D spatial misalignment. The resulting proposals are then refined in the second stage with an auxiliary task of 3D point reconstruction. An effective multi-agent collaborative data augmentation strategy is further proposed to enrich the training data while minimizing information loss. Extensive experiments on public real-world datasets demonstrate that our CoGoal3D achieves new state-of-the-art performance, with 3D AP@0.7 improvements of 10.86\%, 10.34\%, and 10.18\% on the DAIR-V2X, V2V4Real, and V2X-Real datasets, respectively. 
Code is available at \url{https://github.com/Megalo-f/CoGoal3D}.

  \keywords{Autonomous driving \and Collaborative perception \and 3D object detection}
\end{abstract}

\section{Introduction}
\label{sec:intro}

Environmental perception is a fundamental task in autonomous driving. However, single-vehicle perception
is inherently limited by the restricted sensing range and occlusions, leading to inferior performance at complex scenarios such as road crossings. In recent years, V2X collaborative perception has emerged as a highly attractive solution to address these limitations by leveraging multi-view information through V2X. 

\begin{figure}[t]
	\centering
	\includegraphics[width=0.6\linewidth]{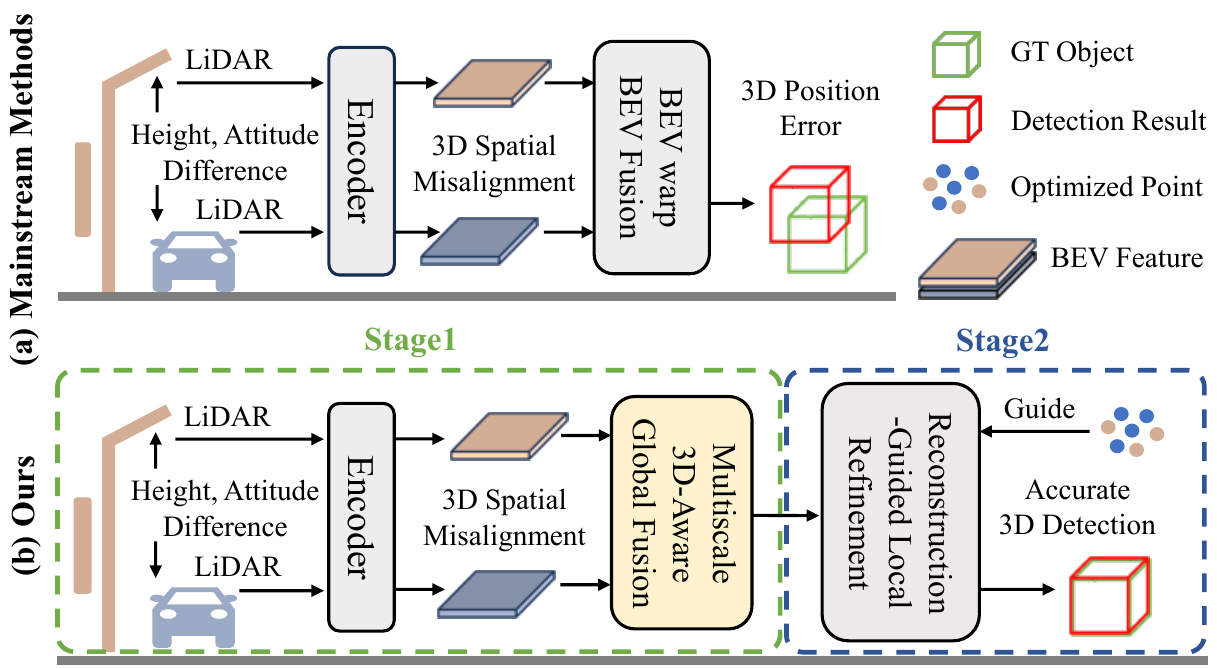}
	\caption{Difference between the mainstream methods and ours. (a) Mainstream broadcast-based methods perform feature fusion only on 2D BEV space. (b) Ours is a two-stage pipeline, enhancing the 3D alignment by 3D-aware global fusion and reconstruction-guided local refinement respectively.
	}
	\label{paradigm}
	
\end{figure}

Currently most of the V2X collaborative perception methods focus on 2D BEV object detection. They usually adopt BEV-based intermediate feature fusion scheme upon a broadcast communication paradigm~\cite{ li2021learning,wang2020v2vnet,xu2022cobevt,hu2022where2comm,lu2023robust}, as shown in Figure~\ref{paradigm}(a). 
In this paradigm, each collaborative agent independently extracts its own BEV feature locally and transmits it to the ego-vehicle together with its pose. The ego-vehicle then aligns the collaborative BEV features to its own coordinate system via a 2D warping operation for subsequent feature fusion. This paradigm is efficient in that it requires only one round of communication and the features extracted for self-perception can be directly broadcasted to others without any extra computational cost. However, the alignment with only 2D BEV-based warping assumes all agents observe the scene in the same horizontal plane, regardless of their discrepancies in height and attitude (e.g., pitch angle) caused by different sensor mounting position as well as uneven ground surface. The problem becomes even more pronounced when 3D instead of 2D BEV object detection task is considered. 

On the other hand, data augmentation is crucial for deep learning based detection methods. However, currently very few data augmentation methods are specially designed for collaborative perception tasks. The popular method, Dual Point Transformation Projection~\cite{xu2024v2x}, projects the point cloud of all collaborative agents into a unified coordinate system (e.g., ego-vehicle's coordinate system) for unified global augmentations, and then reprojects the augmented point cloud back to each agent's respective coordinates before using it for training. Although simple, these ego-centric augmentations risk shifting collaborators' point cloud out of their detection range, thereby causing significant information loss. Better data augmentation method tailored for collaborative perception is highly desired.

In this paper, we propose a novel collaborative 3D object detection framework called CoGoal3D, to address the aforementioned problems. The general pipeline of the network is shown in Figure~\ref{paradigm}(b). In contrast to the mainstream methods, we carefully consider the 3D spatial misalignment problem during BEV fusion, and rely on a two-stage pipeline to gradually align and fuse the collaborative features. Instead of simple 2D warping of BEV features, we propose a multiscale 3D-Aware Global Fusion (3D-AGF) module in stage 1 to embed the 3D position and aggregate the spatial features more robustly. With the object proposals at hand, we further design an auxiliary 3D point reconstruction task in stage 2 to help optimize the 3D bounding box. We also invent a Multi-Agent Collaborative Data Augmentation (MCDA) strategy which is specifically tailored for the collaborative perception. By combining a special sequence of local and global data transformation, MCDA features little information loss and is highly effective in augmenting the collaborative data. We conduct extensive experiments on widely used real-world collaborative perception datasets: DAIR-V2X~\cite{yu2022dair}, V2V4Real~\cite{xu2023v2v4real} and V2X-Real~\cite{xiang2024v2x}. Experimental results show that our method achieves much higher performance than the SOTA methods.

In summary, the main contributions are summarized as follows:
\begin{itemize}
	\item We propose CoGoal3D, a novel collaborative 3D object detection framework that well addresses the 3D spatial misalignment problem underestimated in existing mainstream methods.
	
	\item We design a multiscale 3D-aware global fusion module and an auxiliary 3D point reconstruction task respectively in the two-stage processing pipeline, which gradually refines the feature for the detection.
	
	\item We propose the multi-agent collaborative data augmentation, an effective data augmentation strategy tailored for the collaborative perception task.
	
	\item We conduct comprehensive experiments on public real-world datasets. Experimental results demonstrate that our CoGoal3D outperforms previous SOTA methods on both BEV and 3D AP metrics by a large margin. 
\end{itemize}

\section{Related Work}
\subsection{Single-Vehicle 3D Object Detection}
Single-vehicle 3D object detection can be 
categorized into one-stage
and two-stage
methods based on whether the region proposals are utilized for refinement. In one-stage methods, PointNet~\cite{qi2017pointnet} and PointNet++~\cite{qi2017pointnet++} directly encode irregular raw point cloud to extract point-level features and predict 3D bounding boxes. VoxelNet~\cite{zhou2018voxelnet} divides the raw point cloud into regular voxels and uses 3D convolution to encode voxel features. SECOND~\cite{yan2018second} introduces sparse 3D convolution to accelerate voxel feature encoding. PointPillars~\cite{lang2019pointpillars} partitions the point cloud into regular pillars on the X-Y plane, enabling the use of 2D convolution to encode BEV features to reduce computational consumption.

In two-stage methods, Point-RCNN~\cite{shi2019pointrcnn} uses PointNet++~\cite{qi2017pointnet++} as the backbone to generate proposals and introduces point cloud RoI pooling to extract proposal features. PV-RCNN~\cite{shi2020pv} utilizes both point-based and voxel-based representations of point cloud, extracting proposal features through RoI grid pooling. Voxel-RCNN~\cite{deng2021voxel} uses voxel-based representations to balance detection accuracy and efficiency. Pillar-RCNN~\cite{shi2023pillar} represents point cloud with pillars and extracts proposal features via 2D RoI pooling on the BEV plane.

\subsection{Collaborative 3D Object Detection}
According to the communication paradigm, current collaborative object detection methods can be classified into handshake-based and broadcast-based ones. Handshake-based paradigm requires the ego-vehicle to send its pose at first, allowing other collaborators to project their point cloud to ego-vehicle's coordinate system before extracting the feature and returning the messages. In this category, V2X-ViT~\cite{xu2022v2x} uses a transformer architecture for feature fusion. DI-V2X~\cite{li2024di} introduces domain-mixing instance augmentation and follows DiscoNet~\cite{li2021learning}'s knowledge distillation framework. DSRC~\cite{zhang2025dsrc} utilizes intermediate fusion with augmented point clouds as teacher network for knowledge distillation, enhancing the robustness of the student network. ERMVP~\cite{zhang2024ermvp} improves communication efficiency via feature sampling and handles localization errors with a spatial calibration module. While relatively accurate, handshake-based paradigm suffers from a two-round communication and multiple independent feature extraction for each collaborating agent, resulting in significant communication and computational overhead.

To address these concerns, the broadcast-based paradigm has emerged as the dominant research direction. In this paradigm, the ego-vehicle receives BEV features and poses from other agents and warps the received features to its own coordinate system for subsequent feature fusion. In this line, DiscoNet~\cite{li2021learning} employs early fusion as teacher network for teacher-student knowledge distillation.  CoBEVT~\cite{xu2022cobevt} introduces fused axial attention to capture both local and global relationships, effectively aggregating features across agents. CoAlign~\cite{lu2023robust} proposes a pose-graph optimization method to improve the robustness of collaborative object detection against pose noise. CoSDH~\cite{xu2025cosdh} presents a hybrid intermediate-late fusion paradigm that leverages confidence-aware late fusion to improve robustness against low communication bandwidth.
However, these existing works mainly focus on 2D BEV detection and ignore the differences in height and attitude among collaborators. When the 3D detection task is considered, they obtain inferior performance since critical 3D spatial information can hardly be compensated by simple 2D BEV warping. In contrast, our method fully accounts for the spatial alignment in 3D, by embedding the 3D pose information and supervising the 3D reconstructed points in the pipeline. 

\section{Method}
\label{Method_Sec}
\subsection{Overview}
\begin{figure}[h]
	\centering
	\includegraphics[width=0.9\linewidth]{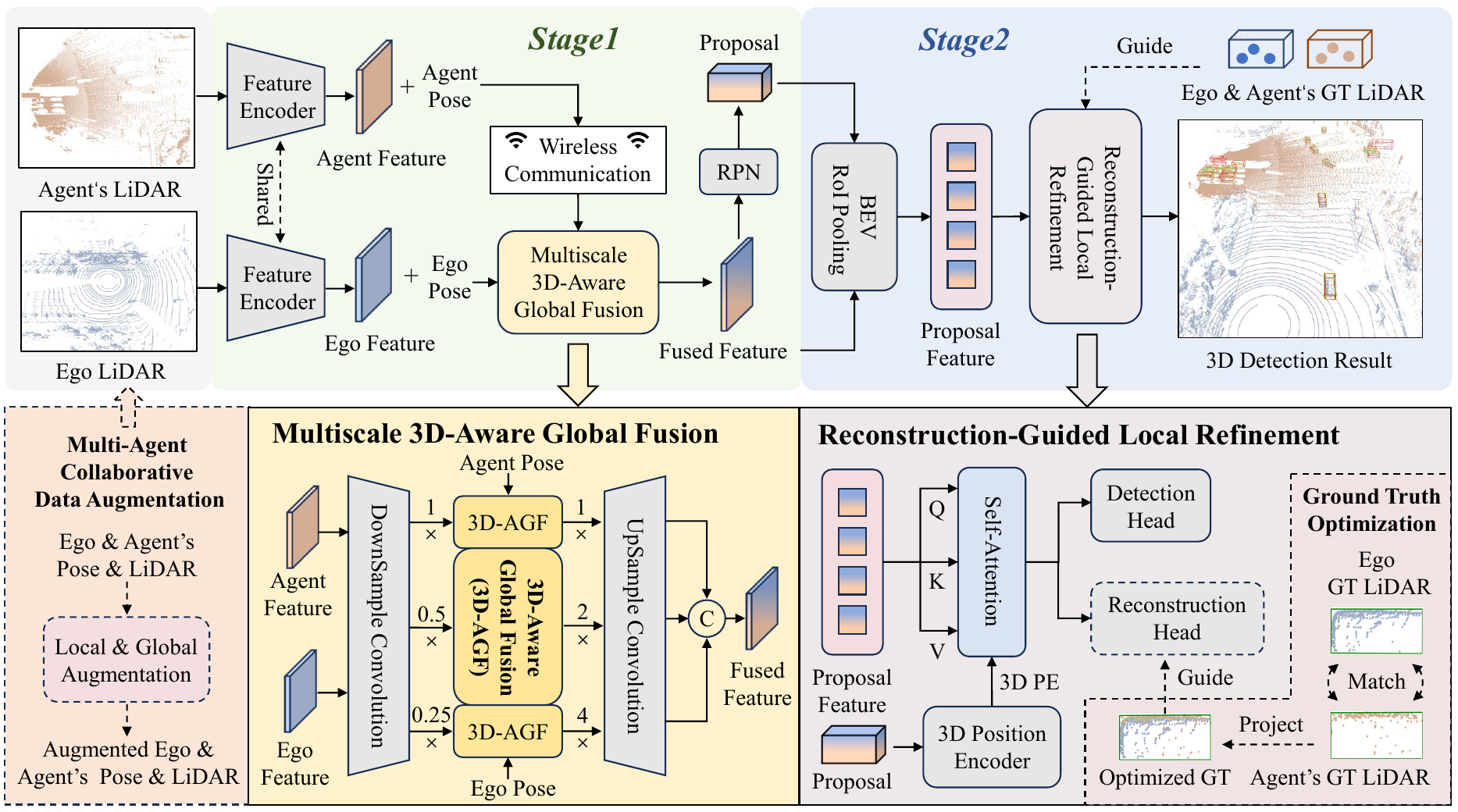}
	\caption{
		The overall architecture of the proposed CoGoal3D. The dashed boxes and lines indicate the components used exclusively during training. Further details of these components are illustrated in Section~\ref{Method_Sec}.
	}
	\label{Overview}
\end{figure}

The overall architecture of the proposed CoGoal3D is illustrated in Figure~\ref{Overview}. It consists of two stages, with multiscale 3D-Aware Global Fusion (3D-AGF) as stage 1 to produce object proposals and Reconstruction-Guided Local Refinement (RGLR) as stage 2 to generate final results. During training, Multi-Agent Collaborative Data Augmentation (MCDA) is applied to the point clouds and poses of all collaborative agents.

In the stage 1,  the input LiDAR points are passed through shared 3D backbones, from which each agent extracts its individual Bird's-Eye-View (BEV) features. The collaborative agents then broadcast their BEV features and poses to the ego-vehicle. Upon receiving these messages, ego-vehicle utilizes the multiscale 3D-Aware Global Fusion (3D-AGF) module to generate the fused feature. Subsequently, a Region Proposal Network (RPN) is used to obtain a set of initial object proposals for the second stage. 

In the second stage, BEV RoI pooling is applied to the fused feature to extract the proposal feature. Then, the obtained feature is fed into the Reconstruction-Guided Local Refinement (RGLR) module for final prediction. It contains two parallel heads: a detection head that produces the refined proposal, and another auxiliary reconstruction head that predicts the 3D point cloud within the proposal. The reconstruction process is supervised by an optimized ground truth point cloud generated by our Ground Truth Optimization (GTO) method, guiding the network to learn fine-grained 3D geometric details of the object, yielding more accurate 3D object results.

Note that MCDA and auxiliary point cloud reconstruction modules are only applied during training. 

\subsection{Multiscale 3D-Aware Global Fusion}
\begin{figure}[h]
	\centering
	\begin{minipage}[b]{0.45\linewidth}
		\centering
		\includegraphics[width=\linewidth]{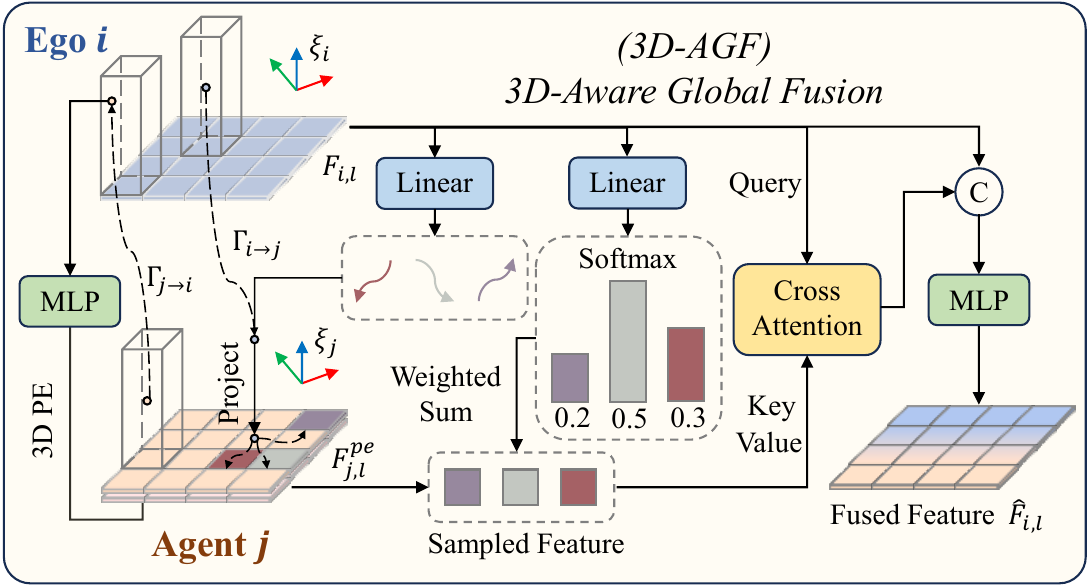}
		\caption{The architecture of the proposed 3D-Aware Global Fusion (3D-AGF) module. It explicitly incorporates 3D spatial information to align collaborative features via 3D position encoding and 3D-aware deformable cross attention.}
		\label{3D-FF}
	\end{minipage}
	\hfill
	\begin{minipage}[b]{0.45\linewidth}
		\centering
		\includegraphics[width=\linewidth]{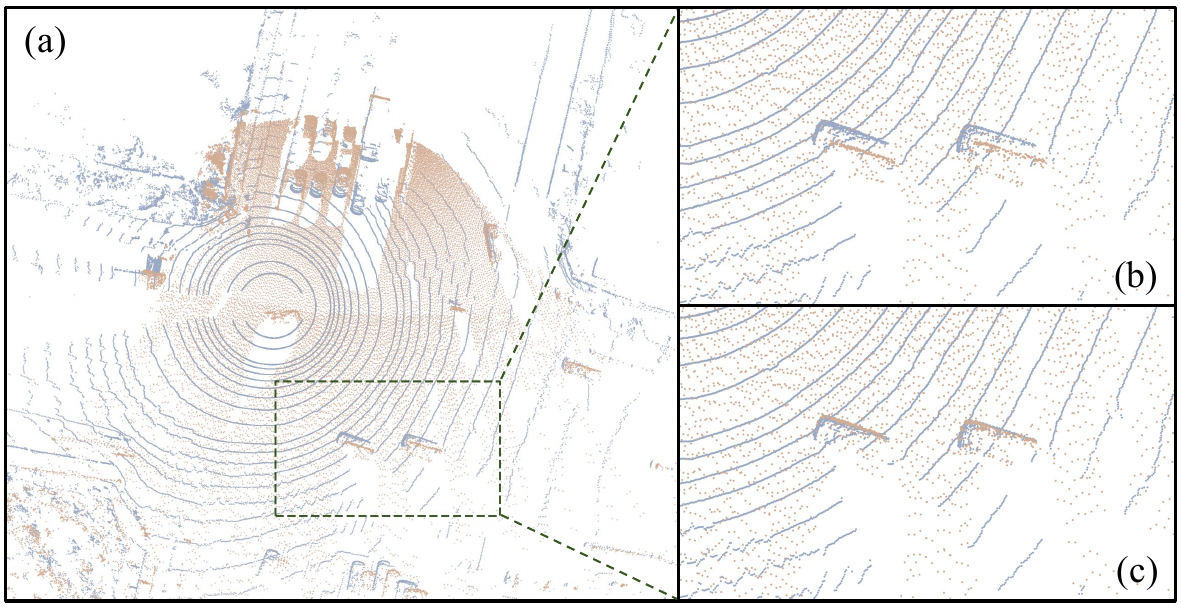}
		\caption{Illustration of Ground Truth Optimization (GTO). (a) is nominally aligned raw point clouds derived from DAIR-V2X dataset. (b) presents an enlarged visualization of the green region in (a). (c) shows the optimized result of the green region in (a) with our GTO.}
		\label{Refine_vis}
	\end{minipage}
\end{figure}

Existing broadcast based methods simply warp the received BEV features on the BEV plane for subsequent feature fusion, which neglects the 3D spatial misalignment caused by different height and attitude among collaborators. To address this, we propose a multiscale 3D-Aware Global Fusion (3D-AGF) module to achieve 3D spatial alignment. We first encode the BEV feature into multiscale features $ {F}_{k,l}, l\in \{1,\cdots,L \}$, where $ {F}_{k,l}$ denotes $k$-th agent's BEV feature at the $l$-th scale, and then performs 3D-AGF at each scale.  Figure~\ref{3D-FF} illustrates the process of 3D-AGF, which consists of the following two steps.

\noindent\textbf{3D Position Encoding.}
To incorporate the crucial 3D spatial information, we introduce 3D position encoding to the BEV features. The relative spatial transformation matrix $ {\Gamma}_{j \rightarrow i}$ is first computed based on the poses of ego-vehicle $ {\xi}_i$ and collaborative agent $ {\xi}_j$. 
Then, we take the corresponding agent's pillar centers in the $l$-th scale as its 3D coordinates $ {p}_{j,l}=(x_{j,l}, y_{j,l}, z_{j,l})$, 
and transform it to the ego-vehicle's coordinate system with $ {\Gamma}_{j \rightarrow i}$ as:
\begin{equation}
	{p}_{j \rightarrow i, l}= {\Gamma}_{j \rightarrow i} *  {p}_{j,l}.
\end{equation}

The obtained $ {p}_{j \rightarrow i, l}$ is passed through an MLP layer to obtain the 3D position encoding, which is then added to $j$-th agent's BEV feature as:
\begin{equation}
	F_{j,l}^{pe} = F_{j,l} + \text{MLP}( {p}_{j \rightarrow i,l}).
\end{equation}

\noindent\textbf{3D-Aware Deformable Cross Attention.} 
Unlike previous methods that first warp collaborative agents' features to the ego-vehicle's coordinate system and then perform feature fusion, we jointly handle 3D global spatial alignment and feature fusion using deformable attention~\cite{zhu2020deformable}. Concretely, we take each grid feature of the ego BEV feature $F_{i,l}$ as a query $q_{i,l}$ and transform its pillar center to $j$-th agent's coordinate system with $\Gamma_{i\rightarrow j}$. The resulting 3D point is then projected to $j$-th agent's BEV plane to obtain the reference point $r_{q,l}$, which compensates for spatial offsets caused by collaborator attitude differences.

With this 3D-aware reference point, deformable attention then learns sampling offsets around $r_{i,l}$ to sample features from $F_{j,l}^{pe}$, yielding the corresponding aligned collaborator's feature $F_{j\to i,l}(q)$ as:
\begin{equation}
	\begin{array}{cc}
		{F}_{j\rightarrow i,l}(q)= \text{Deformable-Attention}(q_{i,l}, r_{q,l}, F_{j,l}^{pe}) = \\
		\sum\limits_{m=1}^{M} W_m {[} \sum\limits_{k=1}^{K} A_{mkq} \cdot (  W'_m F^{pe}_{j,l}(r_{q, l} + \Delta r_{mkq,l}))  {] ,}
	\end{array}     
\end{equation}

where $M$ is the number of attention heads and $K$ is the number of the sampling points. $A_{mkq}$ and $\Delta r_{mkq,l}$ denote the attention weight and sampling offset. $W_m$ and $W'_m$ are the learnable matrices, respectively.

Next, we apply cross attention between the aligned feature ${F}_{j\rightarrow i,l}$ and the ego feature $F_{i,l}$. The result is then concatenated with ego feature $F_{i,l}$ and fused by an MLP as:
\begin{equation}
	\hat{F}_{i,l} = \text{MLP}[F_{i,l}, \text{Cross-Attention}(F_{i,l}, {F}_{j\rightarrow i,l})] .
\end{equation}

Here, $\hat{F}_{i,l}$ denotes the fused feature for the ego at the $l$-th scale, which will be further aggregated into the multiscale fused feature $\hat{F}_i$ by upsample convolution and concatenation, as shown in Figure~\ref{Overview}. Finally, a decoder is used to decode the fused feature $\hat{F}_i$ into a set of proposals $B$.

\subsection{Reconstruction-Guided Local Refinement}
Building upon the global feature fusion provided by multiscale 3D-AGF, we extend our model to a two-stage fusion framework for fine-grained 3D refinement. As shown in Figure~\ref{Overview}, besides a detection decoder for 3D detection task, in this stage we further introduce an auxiliary 3D point reconstruction task to enhance the model's comprehension of 3D spatial information.
It consists of the following two key components.

\noindent\textbf{RoI-level 3D Point Reconstruction.} 
Given the fused feature $\hat{F}_i$ and the proposal $b_m \in B$, we perform BEV RoI pooling~\cite{shi2023pillar} to obtain the proposal features $\{F_{b_m}^{g}\}_{g=1,\cdots,G^2}$, where $G^2$ denotes $G\times G$ regular pillars on the BEV plane that the proposal is divided into. 
Then, 3D position encoding is generated for the proposal features. Specifically, for the $g$-th pillar of the proposal, we denote $p_g$ as the pillar center. The relative coordinates of the pillar center with respect to the proposal's center and vertices are encoded by an MLP as:
\begin{equation}
	p_g^{pos} = \text{MLP}([p_g - r_c;p_g-r_1;\cdots;p_g-r_8]) ,
\end{equation}

where $r_c$ and $r_i$ represent the center and the $i$-th vertex of the proposal. The proposal features are then added with their corresponding position encoding and passed through a self-attention layer, which enables sufficient interaction between the proposal features. The resulting enhanced proposal feature $F_{b_m}$ is represented as:
\begin{equation}
	F_{b_m} = \text{Concat}(\text{SA}(\{F_{b_m}^{g} + p_g^{pos}\}_{g=1,\cdots,G^2})) ,
\end{equation}

where SA and Concat denote the self-attention and concatenating operation. Finally, a decoder is applied to reconstruct a fixed number $N_p$ of points for each pillar of the proposal.

\noindent\textbf{Ground Truth Optimization.} 
Directly using the nominally aligned raw point clouds from collaborators as the ground truth of the reconstruction suffers from slight spatial inconsistencies caused by some artifacts (e.g., calibration errors, temporal asynchrony or LiDAR scanning effects), as shown in Figure~\ref{Refine_vis}(b). We propose a Ground Truth Optimization (GTO) method to provide better supervision signal for the stage 2.  

Specifically, we match the ground truth bboxes of each collaborator with those of the ego-vehicle. For datasets without ID annotations, we match the gt bboxes based on their IoU using the Hungarian algorithm~\cite{kuhn1955hungarian}. Otherwise, we match the gt bboxes directly based on the annotated IDs. For each matched GT pair, we compute the relative transformation between them. This computed relative transformation is subsequently applied to project the collaborator's points within the GT bbox to the matched bbox of the ego-vehicle's. As shown in Figure~\ref{Refine_vis}(c), our GTO yields a better aligned ground truth for supervision, which in turn guides the network to learn fine-grained 3D geometric details, leading to more accurate 3D object detection result.

\subsection{Multi-Agent Collaborative Data Augmentation}

\begin{figure}[h]
	\centering
	\begin{minipage}[b]{0.45\linewidth}
		\centering
		\includegraphics[width=\linewidth]{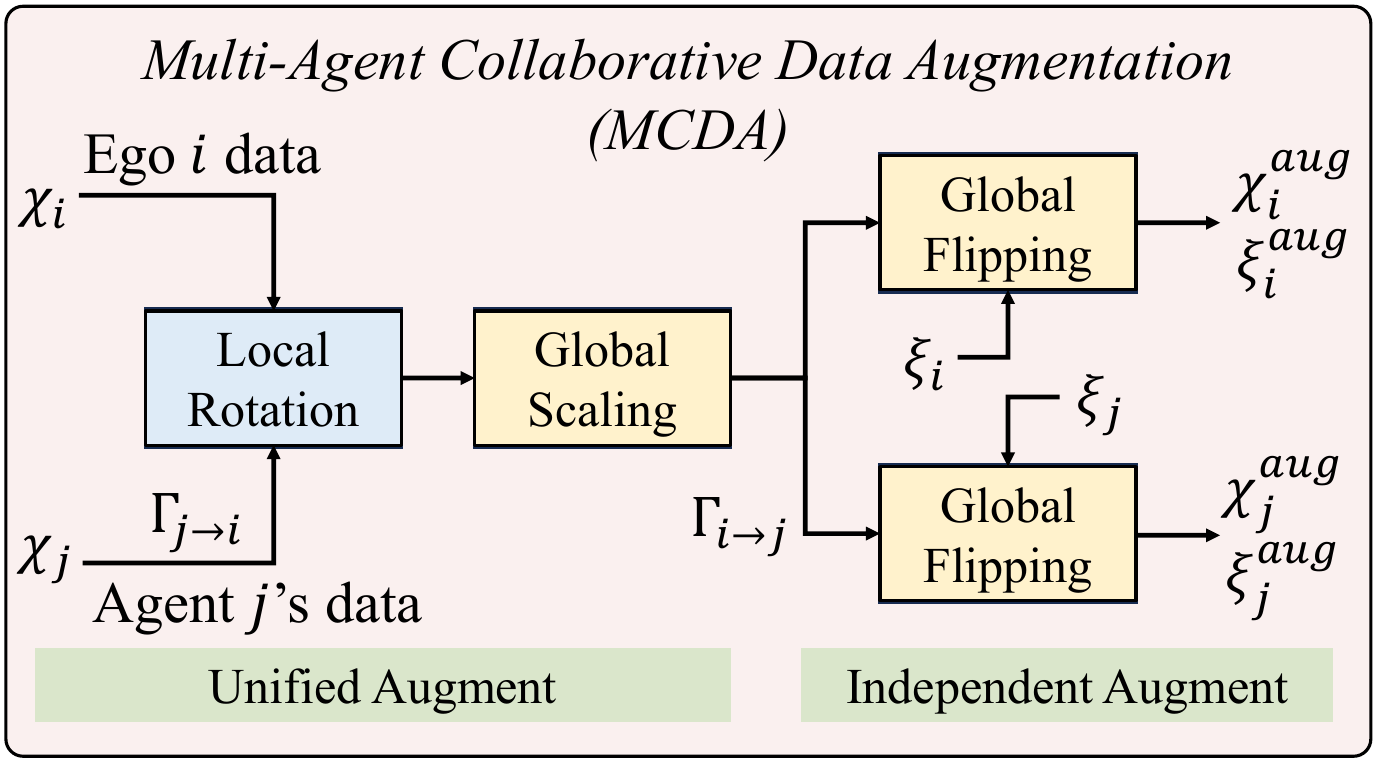}
		\caption{Illustration of the proposed Multi-Agent Collaborative Data Augmentation (MCDA) strategy. It applies a specific sequence of local and global augmentations, maximizing data diversity while minimizing information loss.}
		\label{MTDA}
	\end{minipage}
	\hfill
	\begin{minipage}[b]{0.45\linewidth}
		\centering
		\includegraphics[width=\linewidth]{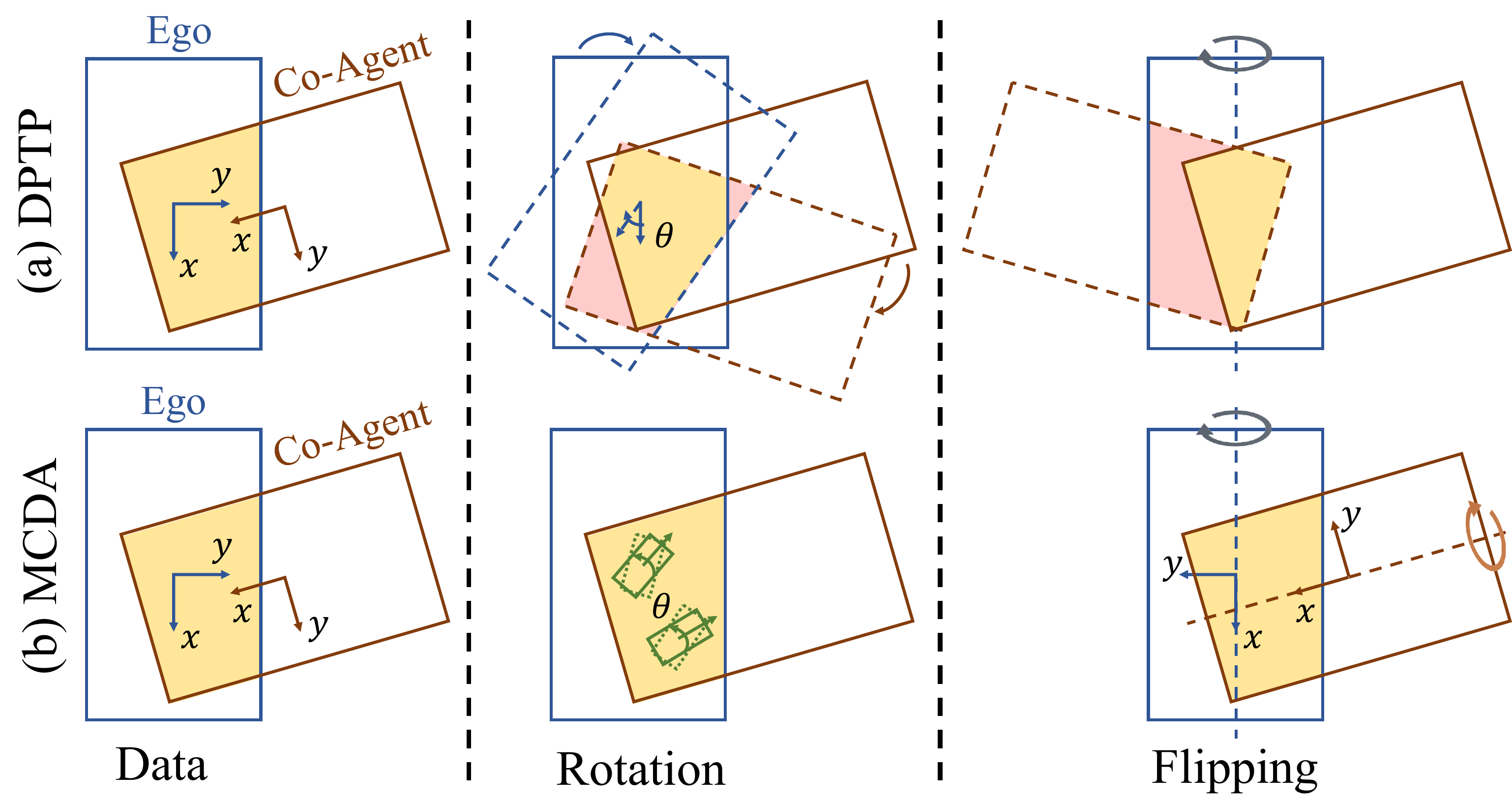}
		\caption{Comparison between DPTP and MCDA. The detection range for the agents are marked with the solid box. The collaborative overlap region and resulting information loss are highlighted with yellow and pink.}
		\label{DataAug_Vis_Comparison}
	\end{minipage}
\end{figure}

Previous data augmentation methods for collaborative perception, such as Dual Point Transformation Projection~\cite{xu2024v2x}, are similar to those for single vehicle perception except for introducing an extra coordinate transformation between the ego and the collaborator. They perform global transformation-based augmentations (e.g., flipping, rotation and scaling) in a unified ego-vehicle's coordinate system. However, such ego-centric global transformations are likely to shift the collaborators' point clouds out of their original detection range, leading to significant information loss, as shown in Figure~\ref{DataAug_Vis_Comparison}(a). To address this issue, we propose Multi-Agent Collaborative Data Augmentation (MCDA), which combines a special sequence of local and global augmentations to maximize data diversity while minimizing the information loss, as shown in Figure~\ref{MTDA}.

Specifically, MCDA first projects the $j$-th agent's points $\chi_j$ into the ego-vehicle's coordinate system with relative transformation matrix $\Gamma_{j\rightarrow i}$
, followed by a unified local rotation around the ground truth objects instead of global rotation. The local rotation augments the object's orientation without altering its global position, thus avoiding shifting the GT objects out of the detection area. Next, a unified global scaling is applied to expand entire training data with different scales. Then, the augmented points of the collaborator are reprojected back to its own coordinate system. Finally, global flipping is performed for each agent, where both the point clouds and the poses are flipped to ensure the spatial consistency. Since the detection range is symmetric, this flipping does not lead to any information loss. As shown in Figure~\ref{DataAug_Vis_Comparison} (b), our MCDA achieves diversified data augmentation with negligible information loss compared to DPTP, which is valuable for the training of collaborative perception tasks.

\subsection{Training Loss}
CoGoal3D adopts an end-to-end training strategy, with the training loss consisting of three components: RPN loss $L_{\text{RPN}}$, proposal refinement loss $L_{\text{refine}}$, and 3D point reconstruction loss $L_{\text{rec}}$, as follows:
\begin{equation}
	L_{\text{total}} = L_{\text{RPN}} + L_{\text{refine}} + L_{\text{rec}},
\end{equation}

$L_{\text{RPN}}$ and $L_{\text{refine}}$ are composed of classification loss and regression loss, corresponding to the first-stage and second-stage detection loss, respectively. To address the class imbalance in the first-stage, the classification loss of $L_{\text{RPN}}$ uses Focal Loss~\cite{lin2017focal}, while $L_{\text{refine}}$ employs binary cross-entropy loss. Smooth L1 loss is applied as the regression loss of both $L_{\text{RPN}}$ and $L_{\text{refine}}$. The 3D point reconstruction loss $L_{\text{rec}}$ is computed by calculating the Chamfer Distance between the reconstructed point cloud and the optimized mixed ground truth point cloud.

\section{Experiment}

\subsection{Datasets and Evaluation Metrics}

\textbf{Datasets.} We evaluated our method on widely used real-world collaborative perception datasets: DAIR-V2X~\cite{yu2022dair}, V2V4Real~\cite{xu2023v2v4real} and V2X-Real~\cite{xiang2024v2x}.
{DAIR-V2X} is the first real-world V2I collaborative dataset, featuring one vehicle and one collaborative infrastructure. The vehicle is equipped with a 40-line LiDAR, while the infrastructure is equipped with a 300-line LiDAR. The dataset consists of 9k frames of collaborative point clouds, split into training, validation, and test sets with a ratio of 5:2:3. We perform experiments using the annotations completed by CoAlign~\cite{lu2023robust}, with the detection range set to $x\in[-100.8m,100.8m], y\in[-40m,40m], z\in[-3.5m,1.5m]$.
{V2V4Real} is the first real-world V2V collaborative dataset, containing two collaborative vehicles, each equipped with a 32-line LiDAR. The dataset includes 20k frames of collaborative point clouds, split into training, validation, and test sets with the proportions 14,210/2,000/3,986. The detection range is set to $x\in[-140.8m,140.8m]$, $y\in[-38.4m,38.4m]$, $z\in[-5m,3m]$.
{V2X-Real} is a large-scale real-world V2X collaborative dataset, featuring a mixture of multiple vehicles and infrastructures equipped with 128 line LiDARs. The dataset contains 33k LiDAR frames, split into training, validation, and test sets with a ratio of 23379/2270/6850. It supports evaluation from different perspectives, including Vehicle-Centric (VC) and Infrastructure-Centric (IC). For our experiments, we focus on the Vehicle-Centric (VC) setting and evaluate the performance on the car class, with the detection range set to $x\in[-102.4m,102.4m], y\in[-38.4m,38.4m], z\in[-5m,3m]$.

\noindent \textbf{Evaluation Metrics.}
We use both BEV and 3D Average Precision (AP) at Intersection-over-Union (IoU) thresholds of 0.5 and 0.7 to evaluate the performance of the collaborative 3D object detection.

\subsection{Implementation Details} 

Our model is implemented in PyTorch~\cite{paszke2019pytorch} and trained on NVIDIA RTX 3090 GPU. We use the Adam optimizer~\cite{kingma2014adam} with an initial learning rate of 0.001, which is decayed by a factor of 0.1 at epochs 10, 20, and 40. Our model is trained for a maximum of 60 epochs with a batch size of 6. Early stopping is employed to select the best epoch.

For the network architecture, we use PointPillars~\cite{lang2019pointpillars} as the 3D backbone and extract BEV features with a pillar size of 0.4m × 0.4m. The multiscale 3D-AGF module employs a 3-layer multiscale deformable attention with 8 attention heads and 9 sampling points. In the second stage, both the BEV RoI pooling size $G$ and the number of reconstructed points for each pillar of the proposal $N_p$ are set to 6.

During training, our model is augmented by the proposed Multi-Agent Collaborative Data Augmentation (MCDA). Specifically, this includes a  global random flipping along the x-axis with a 50\% probability, a local random rotation with an angle sampled uniformly from $[-\pi/20, +\pi/20]$, 
and a global random scaling with a factor sampled uniformly from $[0.95, 1.05]$. The models used for comparison are augmented with Dual Point Transformation Projection (DPTP)~\cite{xu2024v2x}, which consists of a global random flipping along the x-axis with a 50\% probability, a global random rotation with an angle sampled uniformly from $[-\pi/4, +\pi/4]$, 
and a global random scaling with a factor sampled uniformly from $[0.95, 1.05]$. 

\begin{table}[h]
	\centering
	\caption{Performance comparison with state-of-the-art methods on DAIR-V2X \textit{val} set. The best results are presented in \textbf{bold}, while the second-best results are \underline{underlined}. B denotes broadcast communication and H denotes handshake communication. 
	}
	\label{Overall_Performance_on_DAIR-V2X}
	\begin{scriptsize}
		\begin{tabular}{@{}cccccc@{}} 
			\toprule
			Method & Publication & Comm & BEV AP@0.5/0.7 & 3D AP@0.5/0.7 & FPS \\ 
			\midrule
			
			No Fusion~\cite{lang2019pointpillars} & CVPR 2019 & - & 65.56/53.89 & 59.65/29.44 & \textbf{32.4} \\
			DiscoNet~\cite{li2021learning} & NeurIPS 2021 & B & 73.52/58.15 & 64.01/32.34 & \underline{29.7} \\
			V2X-ViT~\cite{xu2022v2x} & ECCV 2022 & H & 76.23/58.76 & 68.68/33.17 & 16.1 \\
			CoBEVT~\cite{xu2022cobevt} & CoRL 2022 & B & 72.77/57.91 & 64.90/35.55 & 13.6 \\
			CoAlign~\cite{lu2023robust} & ICRA 2023 & B & 78.14/64.81 & 68.80/39.69 & 29.1 \\
			DI-V2X~\cite{li2024di} & AAAI 2024 & H & 79.39/65.39 & 72.54/39.24 & 22.3 \\
			ERMVP~\cite{zhang2024ermvp} & CVPR 2024 & H & 75.37/61.49 & 68.61/37.51 & 12.9 \\ 
			DSRC~\cite{zhang2025dsrc} & AAAI 2025 & H & 74.96/60.23 & 67.95/36.08 & 26.1 \\
			CoSDH~\cite{xu2025cosdh} & CVPR 2025 & B & 78.38/64.84 & 67.95/36.78 & 6.7 \\ 
			\midrule 
			CoGoal3D(Stage1) & - & B & \underline{79.49/67.33} & \underline{73.24/42.66} & 24.8 \\
			\textbf{CoGoal3D(Ours)} & - & B & \textbf{81.75/72.16} & \textbf{76.59/50.55} & 16.8 \\
			\bottomrule 
		\end{tabular}
	\end{scriptsize}
\end{table}
\begin{table}[h]
	\centering
	\caption{Performance comparison with state-of-the-art methods on V2V4Real and V2X-Real \textit{test} sets.}
	\label{Overall_Performance_on_Datasets}
	\begin{scriptsize}
		\begin{tabular}{@{}ccccc@{}} %
			\toprule
			\multirow{2}{*}{Method} & \multicolumn{2}{c}{BEV AP@0.5/0.7} & \multicolumn{2}{c}{3D AP@0.5/0.7} \\ 
			\cmidrule(lr){2-3} \cmidrule(l){4-5} %
			& V2V4Real & V2X-Real & V2V4Real & V2X-Real \\ 
			\midrule
			
			No Fusion~\cite{lang2019pointpillars}   & 55.16/40.65 & 64.58/54.02 & 49.32/\underline{21.38} & 61.54/32.40 \\
			DiscoNet~\cite{li2021learning}          & 75.74/44.96 & 76.60/62.66 & 42.66/14.31             & 69.67/36.95 \\
			V2X-ViT~\cite{xu2022v2x}                & 71.98/48.32 & 79.13/63.23 & 59.61/19.60             & 75.27/40.66 \\
			CoBEVT~\cite{xu2022cobevt}              & 71.00/40.46 & 79.31/65.16 & 46.55/13.22             & 75.90/43.21 \\
			CoAlign~\cite{lu2023robust}             & 76.32/50.90 &  79.64/66.44 & 52.73/17.83             & 75.25/44.30  \\
			ERMVP~\cite{zhang2024ermvp}             & 70.94/41.58 & 79.94/65.79 & 48.38/12.55             & \underline{76.76}/\underline{44.46} \\
			DSRC~\cite{zhang2025dsrc}               & 75.53/\underline{53.07} & 78.64/64.64 & \underline{61.20}/21.16 & 75.30/42.67 \\
			CoSDH~\cite{xu2025cosdh}                & \underline{78.98}/51.25 & \underline{84.35}/\underline{70.02} & 50.87/16.67             & 63.47/25.19 \\
			\midrule
			\textbf{CoGoal3D(Ours)}                 & \textbf{82.68/59.72} & \textbf{86.85/76.98} & \textbf{71.48/31.50} & \textbf{81.72/54.64} \\
			\bottomrule 
		\end{tabular}
	\end{scriptsize}
\end{table}

\subsection{Quantitative Evaluation}
\noindent\textbf{Comparison of Detection Performance.}
We compare our method with the existing state-of-the-art (SOTA) collaborative object detection methods.

The results on the DAIR-V2X dataset are shown in Table~\ref{Overall_Performance_on_DAIR-V2X}. Consistent with existing methods, we report the algorithm's evaluation results on the validation set. Experimental results show that our method ranks the first and outperforms the existing best DI-V2X~\cite{li2024di} by a large margin, i.e., 11.31\% and 6.77\% improvements on 3D and BEV AP@0.7, respectively. It is worth noting that DI-V2X is a handshake-based method, which removes the 3D feature misalignment by performing prior 3D coordinate transformation at the collaborator side. However, it cannot well tackle the residual spatial inconsistencies contained in real-world data, resulting in inferior detection results. In contrast, being a broadcast-based method, our CoGoal3D effectively handles the spatial feature misalignment by the two-stage gradual refinement paradigm, and achieve much better performance in both BEV and 3D AP metrics. This is thanks to the multiscale 3D-aware feature fusion module and the auxiliary 3D point reconstruction task, as well as the effective collaborative data augmentation strategy. Meanwhile, we see that our method with stage1-only also performs better than the existing methods, demonstrating the effectiveness of the multiscale 3D-aware global fusion module.

In terms of efficiency, our first-stage model is remarkably fast at 24.8 FPS as a broadcast-based method. The full model CoGoal3D, while slower, achieves the highest AP and still runs at a real-time level of 16.8 FPS.

The experimental results on V2V4Real and V2X-Real datasets are presented in Table~\ref{Overall_Performance_on_Datasets}, where similar phenomena can be observed. Our method achieves great improvements on these datasets, outperforming DSRC~\cite{zhang2025dsrc} in BEV and 3D AP@0.7 by 6.65\% and 10.34\% on V2V4Real, and by 12.34\% and 11.97\% on V2X-Real, respectively, demonstrating its effectiveness under complex and multi-agent scenarios.

\begin{figure}[h]
	\centering
	\begin{minipage}{0.45\textwidth}
		\centering
		\includegraphics[width=0.9\linewidth]{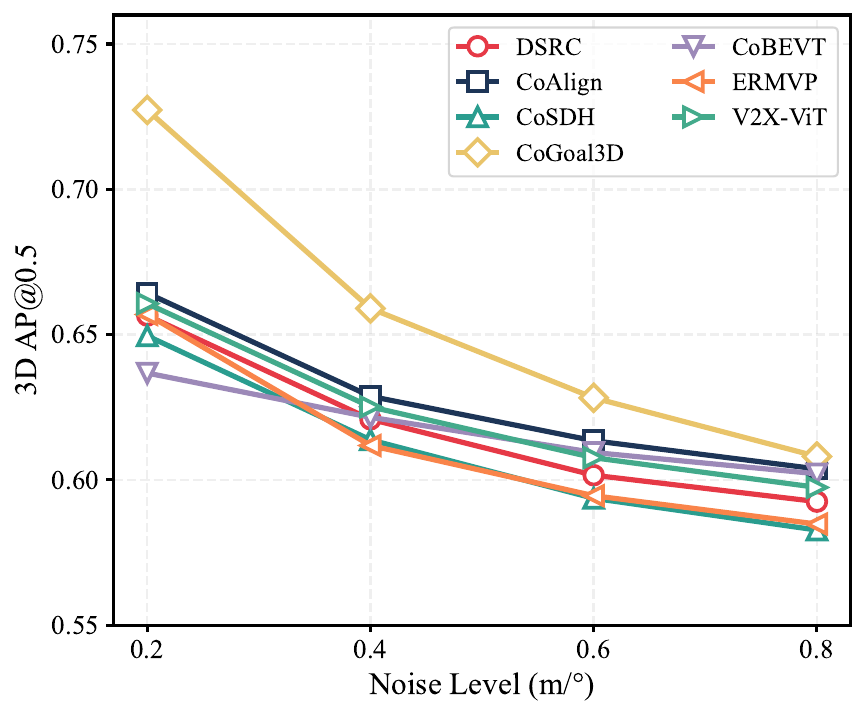}
	\end{minipage}
	\begin{minipage}{0.45\textwidth}
		\centering
		\includegraphics[width=0.9\linewidth]{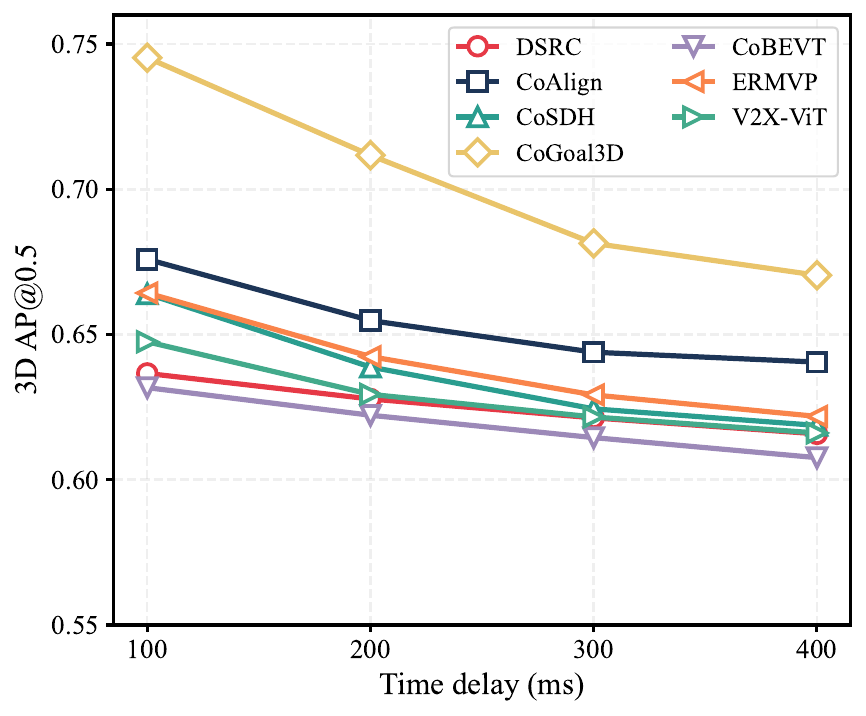}
	\end{minipage}
	
	\caption{Robustness evaluation against localization error (left) and transmission latency (right) on the DAIR-V2X \textit{val} set.}
	\label{fig:robustness_eval}
\end{figure}

\noindent\textbf{Robust Evaluation.}
We further evaluate the robustness of CoGoal3D against localization error and transmission latency. As shown in Figure~\ref{fig:robustness_eval}, we assess the robustness to localization errors by injecting Gaussian noise with varying standard deviations, and simulate transmission latency by introducing different time delays. Thanks to the multi-scale deformable attention that dynamically compensates for spatial misalignment and the optimized point-guided refinement that recovers fine-grained geometric details, CoGoal3D consistently outperforms other methods under these noisy conditions, demonstrating its robustness for practical V2X scenarios.

\subsection{Ablation Study}

We conduct our ablation study on the DAIR-V2X \textit{val} set.

\begin{table}[h] 
	\centering
	
	\begin{minipage}[t]{0.56\linewidth}
		\centering
		\caption{Ablation studies of core components on DAIR-V2X \textit{val} set. MCDA: Multi-Agent Collaborative Data Augmentation; 3D-AGF: multiscale 3D-Aware Global Fusion; RCNN: Second-Stage Refinement; RGLR: Reconstruction-Guided Local Refinement.}
		\label{Ablation_Study}
		\begin{scriptsize}
			\begin{tabular}{@{}ccccc@{}} 
				\toprule
				MCDA  & 3D-AGF & RCNN  & RGLR   & 3D AP@0.5/0.7  \\ 
				\midrule
				&     &  &  & 65.98/33.27 \\ 
				\checkmark   &     &  &  &      71.04/40.85      \\ 
				\checkmark  & \checkmark & &   & 73.24/42.66  \\
				\checkmark  & \checkmark & \checkmark &   & 75.10/48.28   \\
				\checkmark  & \checkmark & \checkmark & \checkmark &\textbf{76.59}/\textbf{50.55} \\
				\bottomrule
			\end{tabular}
		\end{scriptsize}
	\end{minipage}
	\hfill 
	\begin{minipage}[t]{0.42\linewidth}
		\centering
		\caption{Ablation studies of 3D PE and GTO on DAIR-V2X \textit{val} set. 3D PE: 3D Position Encoding in 3D-AGF; GTO: Ground Truth Optimization in RGLR.}	
		\label{3DPE}
		\begin{scriptsize}
			\begin{tabular}{@{}ccc@{}} 
				\toprule
				3D PE & GTO & 3D AP@0.5/0.7  \\ 
				\midrule
				&   &  \textbf{76.59}/\textbf{50.55} \\ 
				$\times$  &  & 73.12/46.70   \\ 
				&   $\times$   & 75.87/49.57      \\ 
				$ \times$ &  $\times$  & 73.29/46.99  \\ 
				\bottomrule
			\end{tabular}
		\end{scriptsize}
	\end{minipage}
	
\end{table}

\noindent\textbf{Ablation of Core Components.}
Table~\ref{Ablation_Study} presents the ablation results of our CoGoal3D. Firstly, introducing the MCDA strategy for network training can greatly improve the baseline by more than 5.06\% on 3D AP@0.5, revealing the superiority of this new data augmentation method. Adding the 3D-AGF module (row 3 vs. row 2) boosts the 3D AP@0.5 by 2.20\%, validating its capability in handling 3D spatial misalignment of the real data. On the other hand, adding RGLR (row 5 vs. row 4) in the second stage improves 3D AP@0.7 by 2.27\%, confirming the benefit of this module. Finally, the full CoGoal3D model outperforms the baseline by 10.61\%/17.28\% on 3D AP@0.5/0.7, demonstrating the effectiveness of our design. More detailed ablation for the core components within the modules will be shown in the following.

\noindent\textbf{Ablation of 3D PE and GTO.}
Table~\ref{3DPE} ablates the role of 3D Position Encoding (3D-PE) in 3D-AGF module and Ground Truth Optimization (GTO) in the RGLR module. The results show that removing either 3D-PE or GTO from CoGoal3D leads to performance drops (3.85\% and 0.98\% in 3D AP@0.7, respectively), confirming their individual contributions in 3D spatial alignment and high quality 3D point supervision. Meanwhile, removing the 3D-PE causes larger performance degradation than the GTO, indicating the importance of the 3D-PE. Interestingly, further remove GTO on the basis of eliminating 3D-PE (row 4) leads to performance improvement (comparing with row 2), which hints that the functionality of 3D reconstruction guidance is dependent on the high-quality 3D-aware fused feature. 

\begin{table}[h]
	\begin{scriptsize}
		\centering
		\begin{minipage}[t]{0.48\linewidth}
			\centering
			\caption{Ablation and comparison between DPTP and MCDA on DAIR-V2X \textit{val} set. F: Flipping. R: Rotation. S: Scaling.}
			\begin{tabular}{@{}ccccc@{}} 
				\toprule
				\multicolumn{3}{c}{Method} & \multicolumn{2}{c}{3D AP@0.5/0.7} \\
				\cmidrule(l){4-5} 
				F & R & S & DPTP~\cite{xu2024v2x} & MCDA \\
				\midrule
				&  &  & 70.52/33.99       & 70.52/33.99 \\ 
				\checkmark & &  & 71.26/37.97     & 72.57/40.14 \\ 
				\checkmark & \checkmark & & 69.76/37.32  & \textbf{73.29}/41.50 \\ 
				\checkmark & \checkmark & \checkmark & 70.02/38.48 & 73.24/\textbf{42.66} \\
				\bottomrule
			\end{tabular}
			\label{DataAug_Comparison}
		\end{minipage}
		\hfill
		\begin{minipage}[t]{0.48\linewidth}
			\centering
			\caption{Generalization performance comparison between DPTP and MCDA on the DAIR-V2X \textit{val} set.}
			\begin{tabular}{@{}ccc@{}} 
				\toprule
				\multirow{2}{*}{Method} & \multicolumn{2}{c}{3D AP@0.5/0.7} \\
				\cmidrule(l){2-3} 
				& DPTP~\cite{xu2024v2x} & MCDA \\
				\midrule
				DI-V2X~\cite{li2024di} & 72.54/39.24 & 73.54/42.54 \\ 
				ERMVP~\cite{zhang2024ermvp} & 68.61/37.51 & 70.82/38.86 \\ 
				DSRC~\cite{zhang2025dsrc} & 67.95/36.08 & 71.63/40.66 \\
				CoSDH~\cite{xu2025cosdh} & 67.95/36.78 & 71.28/39.98 \\ 
				CoGoal3D & \textbf{74.26/48.52} & \textbf{76.59/50.55} \\ 
				\bottomrule
			\end{tabular}
			\label{mtda_result}
		\end{minipage}
	\end{scriptsize}
\end{table}

\noindent\textbf{Ablation of Data Augmentation.}
Table~\ref{DataAug_Comparison} presents the ablation and comparison of our data augmentation strategy (MCDA) against DPTP, using our first-stage model as the baseline. The results show that our MCDA yields consistent performance gains as more transformations for data augmentation are applied, boosting 3D AP@0.7 by 8.67\% in total. In contrast, much less gains are obtained for the DPTP, with final 3D AP@0.5 even becomes 0.5\% lower than the baseline. Comparing the two methods, the key turning point lies in the introduction of different rotation. With the local rotation instead of the global unified one, as well as different flipping operation, our MCDA is able to increase data diversity without introducing information loss.

Table \ref{mtda_result} further illustrates the generalization performance of our MCDA by applying it to existing SOTA methods and comparing against DPTP. The results show that MCDA consistently improves the performance on all methods over DPTP, exhibiting its generality and superiority for the collaborative perception task. As expected, our method CoGoal3D maintains the best performance among all compared approaches under the same augmentation strategy.

\begin{figure}[h]
	\centering
	\includegraphics[width=0.9\linewidth]{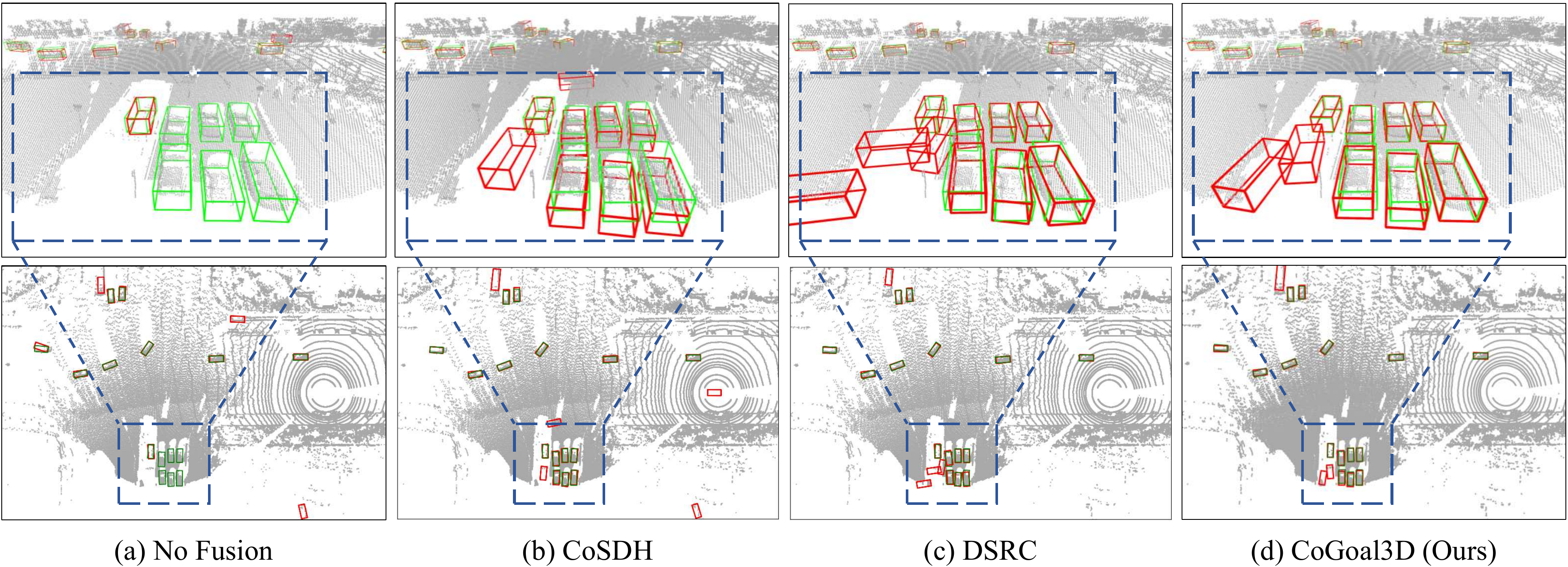}
	\caption{Qualitative results on DAIR-V2X \textit{val} set. The first row shows the 3D zoom-in views of the blue windows in the second row of BEV views. Green and red bounding boxes denote the 3D object ground truths and the detection results, respectively.}
	\label{vis_det}
\end{figure}

\subsection{Qualitative Evaluation}
A representative qualitative comparison between our CoGoal3D and other methods on the DAIR-V2X dataset is shown in Figure~\ref{vis_det}. CoGoal3D obviously achieves better results than others, in that higher consistency of the detected 3D bounding box with the ground truth and fewer false positives. The broadcast-based CoSDH~\cite{xu2025cosdh} exhibits more position errors in the bounding box due to its weakness in correcting 3D spatial misalignment. DSRC~\cite{zhang2025dsrc} has better alignment, but more false positives are predicted due to the inconsistency contained in the real data. This comparison further demonstrates the superiority of our proposed method.

\section{Conclusion}
In this paper, we propose CoGoal3D, a novel collaborative 3D object detection framework, to address the critical issues of 3D spatial misalignment and collaborative data augmentation that are underestimated in existing methods. CoGoal3D performs multiscale 3D-Aware Global Fusion and Reconstruction-Guided Local Refinement in a two-stage manner, gradually refining the 3D feature for the collaborative detection task. A special Multi-Agent Collaborative Data Augmentation strategy tailored for the collaborative task, which features diversifying the training data with little information loss, is also presented. Experimental results on real-world datasets demonstrate that our method achieves superior performance than the existing approaches, validating its effectiveness and great potential in practical application.

\section*{Acknowledgements}
This work was supported by the Key Project of Natural Science Foundation of Zhejiang Province under Grant LZ26F010003, the Key Research \& Development Plan of Zhejiang Province under Grant No.2024C01010, 2024C01017, the Joint R\&D Program of the Yangtze River Delta Community of Sci-Tech Innovation with grant number 2024CSJGG01000, and National Key Laboratory of Collective Intelligence \& Collaboration (Open Fund Project No. QXZ25017101).
%
%
\bibliographystyle{splncs04}
\bibliography{main}
\end{document}